\relax
\documentclass[letterpaper,11pt]{article} 
\usepackage{plain}  
\usepackage{times}  
\usepackage{helvet} 
\usepackage{courier}  
\usepackage[hyphens]{url}  
\usepackage{graphicx} 
\usepackage{natbib}  
\usepackage{caption} 
\usepackage{fullpage}
\usepackage{amsmath}
\usepackage{amssymb}
\usepackage{pifont}
\usepackage{bbding}
\usepackage{booktabs}
\usepackage{multirow}
\usepackage{algorithm}
\usepackage{algorithmic}
\usepackage{makecell}  
\usepackage{natbib}  
\usepackage{threeparttable}
\usepackage{colortbl}
\usepackage[dvipsnames]{xcolor}
\usepackage{tikz}
\usepackage{times}
\usepackage{hyperref}
\usepackage{cleveref}

\usepackage[switch]{lineno}

\newcommand{\ballnumber}[1]{\tikz[baseline=(myanchor.base)] \node[circle,fill=.,inner sep=0.8pt] (myanchor) {\color{-.}\footnotesize #1};}

\font\titlefont=ptmb8t at 18pt
\newcommand{\authorfont}[1]{#1}

\title {\titlefont{{Joint Channel and Weight Pruning for Model Acceleration \\ on Mobile Devices}}}
\date{}

\author{
  \authorfont{
  Tianli Zhao$^{\dagger\mathsection}$, Xi Sheryl Zhang$^\dagger$, Wentao Zhu$^\ddagger$, Jiaxing Wang$^\mathparagraph$, Sen Yang$^\ddagger$, Ji Liu$^\ddagger$, Jian Cheng$^\dagger$} \\ \\
  \authorfont{
  $^\dagger$Institute of Automation, Chinese Academy of Sciences, } \\
  \authorfont{
  $^\mathsection$School of Artificial Intelligence, University of Chinese Academy of Sciences,} \\
  \authorfont{
  $^\ddagger$Kwai Inc., $^\mathparagraph$JD.com} \\
 \texttt{\{zhaotianli2019,xi.zhang\}@ia.ac.cn},\\
 \texttt{\{wentaozhu91,wjiaxing94,senyang.nlpr,ji.liu.uwisc\}@gmail.com},\\
 \texttt{jcheng@nlpr.ia.ac.cn}
}

\begin{document}

\maketitle

\begin{abstract}
    For practical deep neural network design on mobile devices, it is essential to consider the constraints incurred by the computational resources and the inference latency in various applications. Among deep network acceleration related approaches, pruning is a widely adopted practice to balance the computational resource consumption and the accuracy, where unimportant connections can be removed either channel-wisely or randomly with a minimal impact on model accuracy. The channel pruning instantly results in a significant latency reduction, while the random weight pruning is more flexible to balance the latency and accuracy. In this paper, we present a unified framework with \textbf{J}oint \textbf{C}hannel pruning and \textbf{W}eight pruning (JCW), and achieves a better Pareto-frontier between the latency and accuracy than previous model compression approaches. To fully optimize the trade-off between the latency and accuracy, we develop a tailored multi-objective evolutionary algorithm in the JCW framework, which enables one single search to obtain the optimal candidate architectures for various deployment requirements. Extensive experiments demonstrate that the JCW achieves a better trade-off between the latency and accuracy against various state-of-the-art pruning methods on the ImageNet classification dataset. Our codes are available at \url{https://github.com/jcw-anonymous/JCW}.
\end{abstract}

\section{Introduction}
Recently, deep learning has prevailed in many machine learning tasks. However, the substantial computational overhead limits its applications on resource-constrained platforms, e.g., mobile devices. To design a deep network deployable to the aforementioned platforms, it is necessary to consider the constraint incurred by the available computational resource and reduce the inference latency while maximizing the accuracy.

Pruning has been one of the predominant approaches to accelerating large deep neural networks.
The pruning methods can be roughly divided into two categories, channel pruning which removes parameters in a channel-wise manner~\citep{he2017channelprune,zhaung2018discriminationaware,he2018amc}, and weight pruning which prunes parameters randomly~\citep{han2016deepcompression,molchanov17variationaldropout}. The two mainstream pruning methods mainly focus on accelerating neural networks on one single dimension (e.g. either channel wisely or element wisely), while they may have different impacts on the latency and accuracy. For instance, the results in \Cref{tab:joint-motivation} show that the channel pruning method~\citep{guo2020dmcp} offers a better accuracy than the weight pruning method~\citep{elsen2020sparsecov} when the inference latency is high. In contrast, the weight pruning~\citep{elsen2020sparsecov} yields a better accuracy than the channel pruning~\citep{guo2020dmcp} under a low latency requirement. Inspired by this observation, this work attempts to unveil an important problem overlooked before, \emph{is it possible to achieve a better latency-accuracy trade-off by designing a subtle network pruning method that enjoys benefits from both of the two pruning methods?}

\begin{figure}[t]
    \begin{center}
        \includegraphics[width=0.75\linewidth]{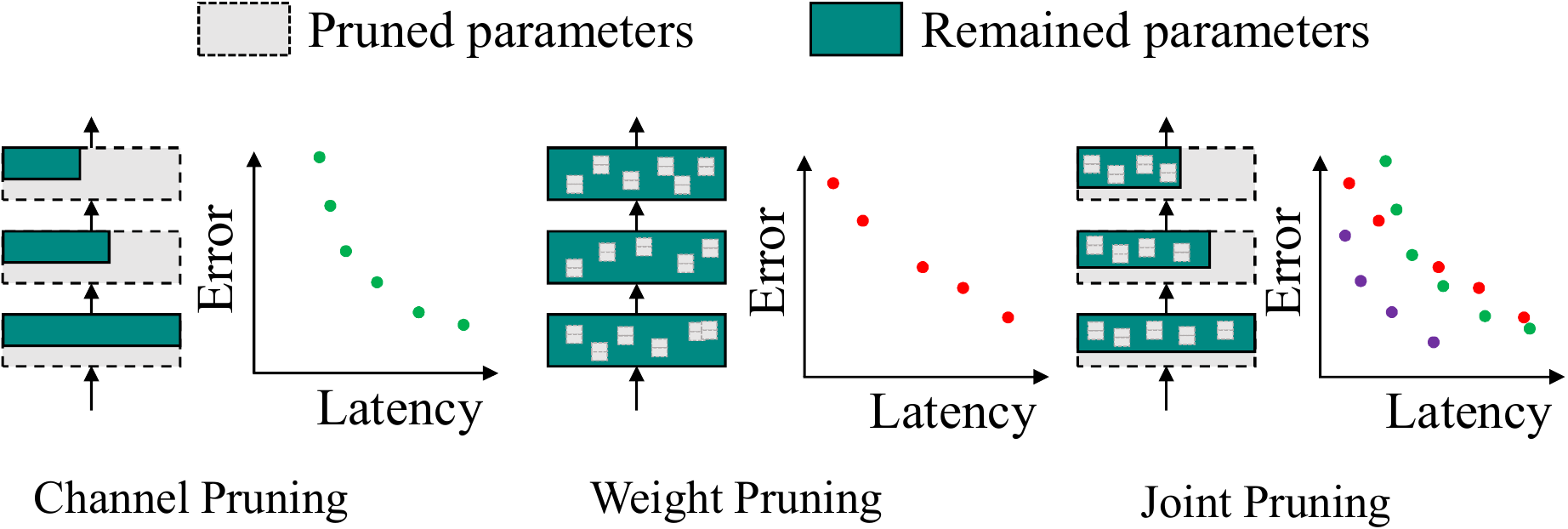}
    \end{center}
    \caption{Illustration of JCW (\textcolor{Plum}{right}), channel pruning (\textcolor{green}{left}) and weight pruning (\textcolor{red}{middle}). The JCW conducts joint channel and weight pruning, which achieves a better Pareto-frontier between the accuracy and latency with one single search.}
    \label{fig:motivation}
\end{figure}

To answer the above question, we leverage both channel pruning and weight pruning to build an efficient model acceleration paradigm, named as JCW (\textbf{J}oint \textbf{C}hannel and \textbf{W}eight pruning). Considering both the latency and accuracy, we formulate the model acceleration into a multi-objective optimization (MOO) problem. Specifically, given a predefined base model with $L$ layers, denoting the number of channels of
certain compressed model for channel pruning by $C = \{c^{(l)}\}_{l=1}^L$, and weight sparsity for weight pruning by $S = \{s^{(l)}\}_{l=1}^{L}$, we search for a sequence of $(C, S)$ lying on the Parato-frontier between latency and accuracy:
\begin{equation}
    (C, S)^* = \arg\min_{C, S}\left\{\mathcal{T}(C, S), \mathcal{E}(C, S)\right\},
\end{equation}
where $\mathcal{T}(C, S)$ and $\mathcal{E}(C, S)$ denote the latency and error rate of the compressed model, respectively. We further propose a uniform non-dominated sorting selection based on an enhanced evolutionary algorithm, NSGA-II~\citep{deb2002nsga2}, to generate optimal candidate architectures with a wide range of latency in one single optimization run. To alleviate the search cost, we construct an accuracy predictor based on parameter sharing~\citep{pham2018enas,guo2020single}, and a latency predictor based on tri-linear interpolation.

We conduct extensive experiments to validate the effectiveness of JCW on the ImageNet dataset~\citep{deng2009imagenet}. The JCW outperforms previous state-of-the-art model compression approaches by a large margin. Without loss of accuracy, the JCW yields $2.74\times$, $2.42\times$ and $1.93\times$ speedup over ResNet18~\citep{he2015resnet}, MobileNetV1~\citep{howard2017mobilenets} and MobileNetV2~\citep{sandler2018mbv2}, respectively.

Our major contributions are summarized as follows,
\begin{itemize}
    \item We build a general model acceleration framework by the joint channel pruning and weight pruning as shown in the right of \Cref{fig:motivation}, which achieves a better trade-off between model accuracy and latency on mobile devices.
    \item We enhance the Pareto multi-objective optimization with a uniform non-dominated sorting selection. The enhanced search algorithm can find multiple optimal candidate architectures for various computational budget through one single search.
    \item Extensive experiments demonstrate the effectiveness of the joint channel pruning and weight pruning. The JCW outperforms previous state-of-the-art model compression and acceleration approaches by a large margin on the ImageNet classification dataset.
\end{itemize}

\section{Methodology}

\subsection{Motivation}
\begin{table}[t]
    \centering
    \setlength\tabcolsep{14pt}
    \begin{tabular}{l|ccc}
    \Xhline{2\arrayrulewidth}
         Method & Type & Latency & Accuracy \\
         \hline

                                Fast & Weight Pruning & 88.94 ms & 72.0\% \\
                              DMCP & Channel Pruning & 82.95 ms & 72.4\% \\
                             \hline
                             Fast & Weight Pruning & 35.89 ms & 65.2\% \\
                             DMCP & Channel Pruning & 33.50 ms & 62.7\% \\
    \Xhline{2\arrayrulewidth}
    \end{tabular}
    \caption{Latency \& accuracy of MobileNetV2 models accelerated by fast sparse convolution~\citep{elsen2020sparsecov} and DMCP~\citep{guo2020dmcp}. The best acceleration strategy differs under various latency budgets.}
    \label{tab:joint-motivation}
\end{table}
In this section, we review the two categories of existing network pruning methods -- structured channel pruning and random weight pruning, and analyze their pros and cons, which give rise to our joint design method.

Channel pruning reduces the width of feature maps by pruning filters channel-wisely~\citep{guo2020dmcp,wang2020revisiting}. As a result, the original network is shrunk into a thinner one. The channel pruning is well-structured and thus conventionally believed to be more convenient for model acceleration than random weight pruning~\citep{wenwei2016SSL,he2017channelprune}. However, a well known drawback of channel pruning is its difficulty in attaining the accuracy because of its strong structural constraints.

In random weight pruning, each individual element of parameters can be freely pruned~\citep{han2015lwc,han2016deepcompression}. It is more flexible and generally known to be able to get theoretically smaller models than channel pruning. More recently, \citet{elsen2020sparsecov} made it more practical by arguing that despite of its irregular memory access, it can also be efficiently accelerated on mobile CPUs if implemented properly. However, the acceleration ratio achieved by pure weight pruning is still limited because the problem of irregular memory access still exists. For example, their method can only accelerate the computation by $\sim 3\times$ even when $90\%$ of parameters are removed.

Based on the above analysis, we raise an important problem overlooked before that: \emph{How do these two categories of pruning methods affect the accuracy and latency of networks? Is it possible to enjoy benefits from both of the two methods by applying random weight pruning and structured channel pruning jointly?} In \Cref{tab:joint-motivation}, we show the comparison of accuracy-latency trade-off between the two acceleration methods, fast sparse convolution~\citep{elsen2020sparsecov} which is one of the state-of-the-art network acceleration methods with weight pruning, and DMCP~\citep{guo2020dmcp} which is one of the state-of-the-art network acceleration methods with channel pruning. We observe that the channel pruning achieves a better accuracy-latency trade-off under a higher latency budget, while weight pruning achieves a better accuracy-latency trade-off under a lower latency budget. This implies that the proper accelerating method may be different under different latency budgets. The fact may be more complicated. Specifically, it is very likely that the optimal choice is to apply the two jointly. This motivates us to investigate \emph{whether it is possible to further achieve a better accuracy-latency trade-off by applying channel pruning and weight pruning simultaneously?}

To answer the above question, we present JCW, a unified framework applying channel pruning and weight pruning jointly for model acceleration. In \Cref{fig:motivation}, we illustrate the comparison between JCW and the two pruning methods. The JCW absorbs the advantages of both methods, thus it is able to achieve a better trade-off between latency and accuracy.

\begin{figure*}[!bt]
    \begin{center}
    \includegraphics[width=0.9 \linewidth]{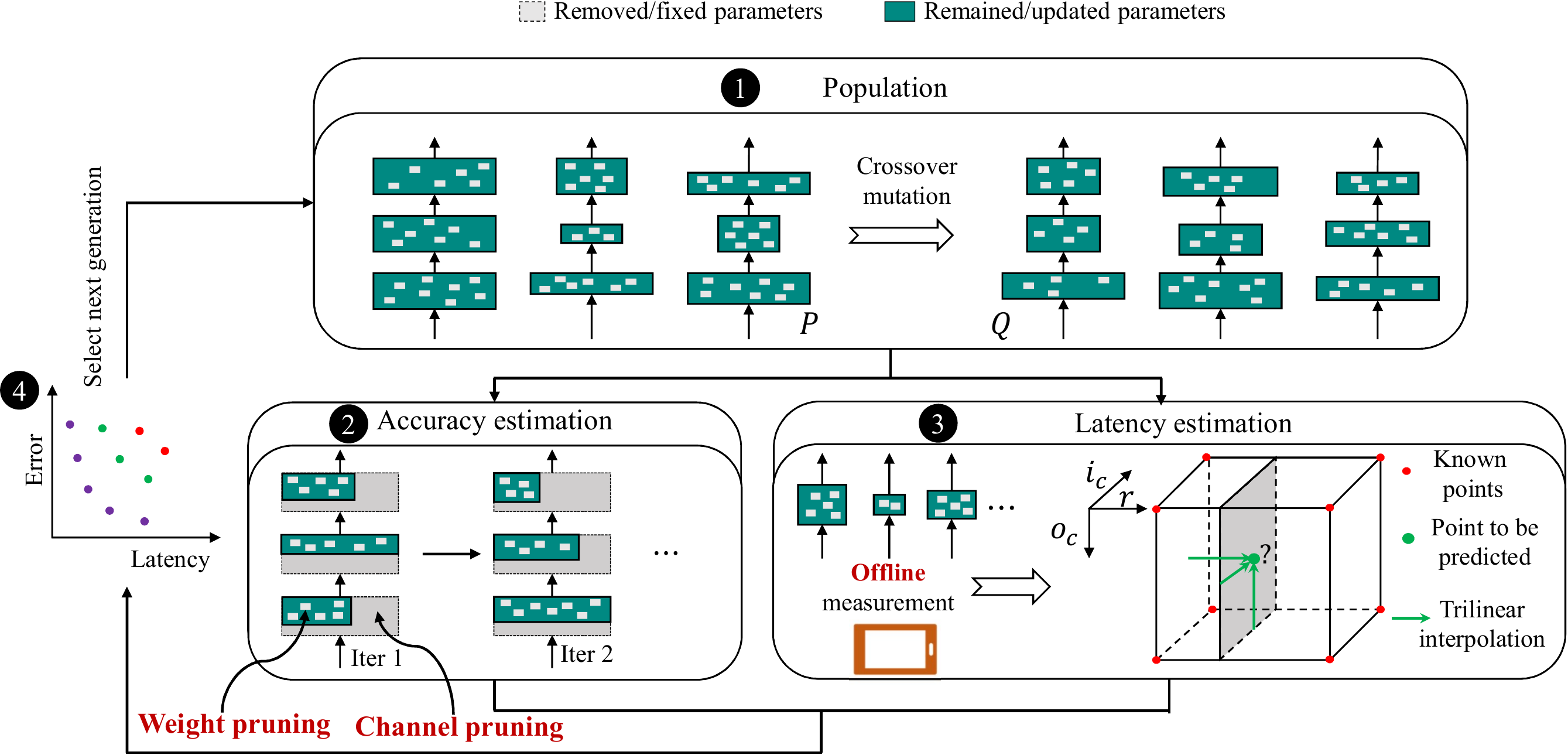}
    \end{center}
    \caption{Illustration of the framework of JCW.}  
    \label{fig:framework}
\end{figure*}
\subsection{Problem formulation}
Formally, for some compressed model $\mathcal{A}$ with $L$ layers, we denote the number of channels of each layer by $C_{\mathcal{A}} = \{c^{(l)}_{\mathcal{A}}\}_{l=1}^{L}$, and the weight sparsity\footnote{We use weight sparsity to denote the ratio of non-zero parameters of remaining channels across the whole paper.} of each layer by $S_{\mathcal{A}} = \{s^{(l)}_{\mathcal{A}}\}_{l=1}^{L}$. In this way, each sub-network $\mathcal{A}$ can be represented by a pair of vectors:
\begin{equation}
    \mathcal{A} = \{C_{\mathcal{A}}, S_{\mathcal{A}}\}.
\end{equation}
Our goal is to accelerate the inference of networks by \emph{applying channel pruning and weight pruning simultaneously}, while at the same time minimizing the accuracy loss:
\begin{equation}
    \begin{split}
        \mathcal{A}^* &= \arg\min_\mathcal{A} \{\mathcal{T}(C_{\mathcal{A}}, S_{\mathcal{A}}), \mathcal{E}(C_{\mathcal{A}}, S_{\mathcal{A}})\},
    \end{split}
    \label{eq:problem-formulation}
\end{equation}
where $\mathcal{T}(C_{\mathcal{A}}, S_{\mathcal{A}})$ and $\mathcal{E}(C_{\mathcal{A}}, S_{\mathcal{A}})$ denote the inference latency and task specific error of the model, respectively. For simplicity, we will abbreviate them as $\mathcal{T}(\mathcal{A})$ and $\mathcal{E}(\mathcal{A})$ in the remaining of this paper under the clear context.

A crucial question for solving problem in \Cref{eq:problem-formulation} is: \emph{How to determine the number of channels and weight sparsity for each layer?} One alternative way is to first prune the channels of the original model in an automated way, then prune the weights of the channel-pruned model for further acceleration. However, this separated optimization may lead to a sub-optimal solution, because it is difficult to find the optimal balance between channel pruning and weight pruning by hand.
Concretely speaking, the optimal architecture for channel pruning may be sub-optimal when further applying weight pruning for acceleration. Therefore, we instead optimize the number of channels and weight sparsity simultaneously in one single optimization run and determine the balance of acceleration between channel pruning and weight pruning automatically.

Another difficulty in solving the problem in \Cref{eq:problem-formulation} is that there are more than one objective (the latency and accuracy) to be optimized, yielding a multi-objective optimization (MOO) problem naturally. It is challenging to design one model that achieves the best values for both of the two objectives, since these two objectives are generally conflict with each other. Therefore, the optimal solutions for problem in \Cref{eq:problem-formulation} are not unique, and we need to find a sequence of models lying on the Pareto-frontier between the accuracy and latency\footnote{In MOO, the Pareto-frontier is a set of solutions that for each solution, it is not possible to further improve some objectives without degrading other objectives.}. We solve this problem based on evolutionary algorithm NSGA-II~\citep{deb2002nsga2}.

\subsection{Unified framework}
Before going into the details, we first introduce the overview of the whole framework, which is illustrated in \Cref{fig:framework}. The JCW works in an iterative way, it maintains a sequence of well-performing models $P = \{(C_i, S_i)\}_{i=1}^{n}$ with different number of channels and weight sparsity, here $n$ is the population size. \ballnumber{1} In each iteration, a new set of models $Q = \{(C_i^{new}, S_{i}^{new})\}_{i=1}^{n}$ are generated from $P$ through crossover and mutation operators. Then, we estimate the accuracy (\ballnumber{2}) and latency (\ballnumber{3}) of all the models in $P\cup Q$. \ballnumber{4} Based on the estimations, we select the models with various latency and relatively low error rate to form the next generation of $P$. The proposed components are integrated under the learning problem of \Cref{eq:problem-formulation}, and the iteration continues until the qualified models are found. In the sequel, we instantiate different components of the framework, i.e. the accuracy estimator in \Cref{sec:acc-estimator}, the latency estimator in \Cref{sec:lat-estimator}, and the uniform non-dominated sorting selection of models in \Cref{sec:selection}.

\subsection{Accuracy estimation}
\label{sec:acc-estimator}
Motivated by~\citep{guo2020single}, we apply the parameter sharing technique for efficient accuracy estimation for compressed models in the JCW so that the joint design can be fulfilled. In particular, suppose $P = \{(C_i, S_i)\}_{i=1}^n$ is a sequence of models whose performances are to be estimated. We create a supernet $\mathcal{N}$ with the maximum number of channels in each layer, and the parameters of the supernet is denoted $W = \{W^{(l)}\}_{l=1}^L$. The overall goal is to train $W$ on the dataset $\mathcal{D}$ to make the accuracy rank of each sub-network in $P$ maintain the same as it is trained independently. We then formulate the problem by:
\begin{equation}
    \min_{W} \mathbb{E}_{(x,y)\sim \mathcal{D}, (C, S)\sim P}[\mathcal{L}(y|x;W_{C, S})],
    \label{eq:acc-problem}
\end{equation}
where $W_{C, S} = W\odot M_{C,S}$, $\odot$ denotes the element-wise multiplication, and $M_{C, S}$ is a stack of binary masks that selects a part of parameters from $W$ to form a sub-network with parametric configurations $(C, S)$. In this paper, we simply select channels with top indices and then apply norm based pruning as a selection scheme for parameter sharing.

In JCW, problem~\eqref{eq:acc-problem} is solved in a batched stochastic gradient descent fashion. Particularly, in each update step, we randomly sample a batch of data $(x, y)$ from $\mathcal{D}$, and a sub-network $(C, S)$ from $P$, calculate the gradient of the loss \textit{w.r.t.} the $W$ by:
\begin{equation}
    g_{C, S} = \frac{\partial L(y|x;W_{C,S})}{\partial W_{C, S}} \odot M_{C, S},
\end{equation}
and update the parameters with the SGD solver.

\subsection{Latency estimation}
\label{sec:lat-estimator}
Previously, latency prediction is conducted by constructing a look up table~\citep{yang2018netadapt,wang2019haq,wang2020apq} or training an estimation model~\citep{yang2019ecc,berman2020aows}. The former one is limited to a small number of candidates, while the latter one requires a large amount of architecture-latency pairs to train the estimation model, which is laborious to collect. In contrast, in the JCW, we propose to estimate the latency of models with trilinear interpolation. The latency of one model can be represented by the summation of the latency of each layer:
\begin{equation}
    \mathcal{T}(C, S) = \sum_{l=1}^{L} \mathcal{T}^{(l)}(C, S),
\end{equation}
where $\mathcal{T}^{(l)}(C, S)$ is the latency of the $l$-th layer of the model represented by $\{C, S\}$. In the context of joint channel pruning and weight pruning, the latency of each layer depends on the number of input/output channels, and the weight sparsity of that layer:
\begin{equation}
    \mathcal{T}^{(l)}(C, S) = \hat{\mathcal{T}}^{(l)}(c^{(l-1)}, c^{(l)}, s^{(l)}).
\end{equation}

For efficient latency estimation, we build a layer-wise latency predictor $\hat{\mathcal{T}}^{(l)}(\cdot)$ with trilinear interpolation. This is based on the observation that the latency is locally linear with respect to the layer width and weight sparsity\footnote{Please refer to the Appendix for more results about this observation.}. Specifically, we denote $C_{max} = \{c^{(l)}_{max}\}_{l=1}^L$ as the maximum number of channels of each layer. For layer $l$ with maximum input channels of $c_{max}^{(l-1)}$ and maximum output channels of $c_{max}^{(l)}$, we first measure the real runtime on the target device with channel width $0, \frac{1}{N}, \frac{2}{N}, \cdots, 1.0$ of both input and output channels, and weight sparsity of $0, \frac{1}{M}, \frac{2}{M}, \cdots, 1.0$, respectively, generating a 3-D array of architecture-latency samples. We denote $T_{ijk}^{(l)}$ to be the latency of the $l^{th}$ layer's convolution with input, output channels and weight sparsity of $\frac{i}{N}c_{max}^{(l-1)}, \frac{j}{N}c_{max}^{(l)}$, and $\frac{k}{M}$, respectively, and define:
\begin{equation}
    c_{i}^{(l)} = N\frac{c^{(l)}}{c_{max}^{(l)}} - i, ~~~~ s_{i}^{(l)} = Ms^{(l)} - i
\end{equation}
to be the normalized array index. Given any input/output channels and weight sparsity, we can easily approximate the latency through the trilinear interpolation of the 3-D array \footnote{More detailed derivation is given in Appendix.}:
\begin{equation}
    \hat{\mathcal{T}}^{(l)}(c^{(l-1)}, c^{(l)}, s^{(l)}) = \sum_{i,j,k} \tau(c_{i}^{(l-1)})\tau(c_{j}^{(l)})\tau(s_{k}^{(l)})T_{ijk}^{(l)},
\end{equation}
where:
\begin{equation}
    \tau(x) = \max(0, 1 - |x|).
\end{equation}

In practice, we find that $M = 10, N = 8$ is sufficient for approximating latency with high efficiency and accuracy as illustrated in \Cref{fig:latency-estimation}. Compared to other latency estimation methods~\citep{yang2018netadapt,berman2020aows} requiring tens of thousands of architecture-latency pairs, the proposed trilinear interpolation based latency predictor can be efficiently constructed with as less as 700 data points.

\subsection{Uniform non-dominated sorting selection}
\label{sec:selection}
To update the well performed individuals in the population, in the original NSGA-II~\citep{deb2002nsga2}, candidates in the combined population are sorted with fast non-dominated sort algorithm after evaluated based on multiple objectives, \textit{i.e.}, the latency and accuracy. In our experiments, we find this standard non-dominated sorting selection tends to prioritize models of low latency as shown in \Cref{fig:latency-dist}. The probable reason is that the accuracy estimation with parameter sharing is not as accurate as latency estimation, and the evolver puts much emphasize on the latency optimization.

To handle this obstacle, we propose a uniform non-dominated sorting selection to generate diverse architectures of various latency. We first uniformly sample $N$ points from the interval $[T_{\min}, T_{\max}]$:
\begin{equation}
    T_i = T_{\min} + i \times \frac{T_{\max} - T_{\min}}{N - 1}, i = 0, 1, 2, \cdots, N - 1,
\end{equation}
where $T_{\min}$, $T_{\max}$ are the minimal and maximal latency, respectively. For each $T_i$, we sort the individuals in the merged population $P\cup Q$ with objectives $\{|T_i - \mathcal{T}|, \mathcal{E}\}$ with non-dominated sort~\citep{deb2002nsga2}, where $\mathcal{T}$ and $\mathcal{E}$ are latency and error rate of the network architecture.

Let $F_{i} = \{F_i^{(s)}\}_{s=0}^{n_i-1}$ be frontier stages after sorting architectures with objectives $\{|T_i - \mathcal{T}|, \mathcal{E}\}$. We select candidates stage by stage in the order $F_{0}^{(0)}, F_{1}^{(0)}, \cdots, F_{0}^{(1)}, F_{1}^{(1)}, \cdots$ until the number of selected candidates reaches the evolutionary population size. When we select individuals from $F_i$, architectures with latency close to $T_i$ and relatively low error rate will be selected. In the right of \Cref{fig:latency-dist}, it demonstrates that our uniform non-dominated sorting selection generates diverse architectures of various latency.

\section{Experimental Results}
To validate the efficiency of our joint channel and weight pruning (JCW), we conduct extensive experiments, including ablation studies, based on ResNet18~\citep{he2015resnet}, MobileNetV1~\citep{howard2017mobilenets}, and MobileNetV2~\citep{sandler2018mbv2} on the ImageNet classification dataset~\citep{deng2009imagenet}. We did not conduct experiments on the CIFAR-10 dataset~\citep{krizhevsky2009learning}, because it is much more challenging and practical to compress and accelerate large deep networks based on a large-scale dataset. We compare our JCW with several state-of-the-art model compression and acceleration approaches and find that the JCW achieves the best trade-off between model accuracy and latency. The ablation study validates the effectiveness of each component in the JCW.


\subsection{Implementation details}
We first present some implementation details of JCW, and more details are provided in Appendix.

\noindent\textbf{Computation of sparse convolution.} We use similar technique as fast sparse ConvNets~\citep{elsen2020sparsecov} for efficient computation of sparse convolution, with two slight improvements. (i) For weight pruning, we group the parameters along the output channels and remove the parameters group-wisely. Specifically, four parameters at the same location of adjacent output channels are grouped together, and parameters in the same group are removed or retained simultaneously. The grouping strategy is beneficial for efficient data reuse~\citep{elsen2020sparsecov}. In all of our experiments, the group size is set to 4. (ii) We extend their computation algorithm to support not only matrix multiplication but also regular convolution. We implement an efficient algorithm for the computation of sparse convolution and utilize it to measure the latency of our searched sparse models.

\noindent\textbf{Evolutionary search.}
For evolutionary search, we set the population size to 64 and the number of search step to 128. We sample a subset of the ImageNet dataset for the supernet training. Specifically, we randomly sample 100 classes from ImageNet, 500 images per class to construct the train set, and randomly sample 50 images from the rest images per class to construct the validation set. We train the supernet for 30 epochs with batch size of 256, where the first 10 epochs are used for parameter sharing warming up. The learning rate and weight decay are set to 0.1 and 0.00004 in all the experiments, respectively. For each model in supernet, the batch normalization (BN) statistics are recalculated with $1,280$ images.

\noindent\textbf{Model re-training.}
After we complete the search stage, the generated architectures are re-trained on the whole training set and validated on validation set of the ImageNet dataset. We train the compressed models with ADMM, details about the hyper parameters of different models are given in Appendix.

\noindent\textbf{Measurement platform.}
We measure the latency of all the models on one single ARM Cortex-A72 CPU. The latency of models for channel pruning methods are measured with TFLite~\citep{google2020tflite}, which is a commonly used mobile-oriented deep learning inference framework for dense deep neural networks. Latency of models with weight sparsity are measured with our implemented high performance sparse convolution algorithm. We run each model for 20 times and report the average of the runtime.

\subsection{Ablation study}

\textbf{Accuracy of latency estimation.}
\begin{figure}[t]
    \begin{center}
        \includegraphics[width=0.9\linewidth]{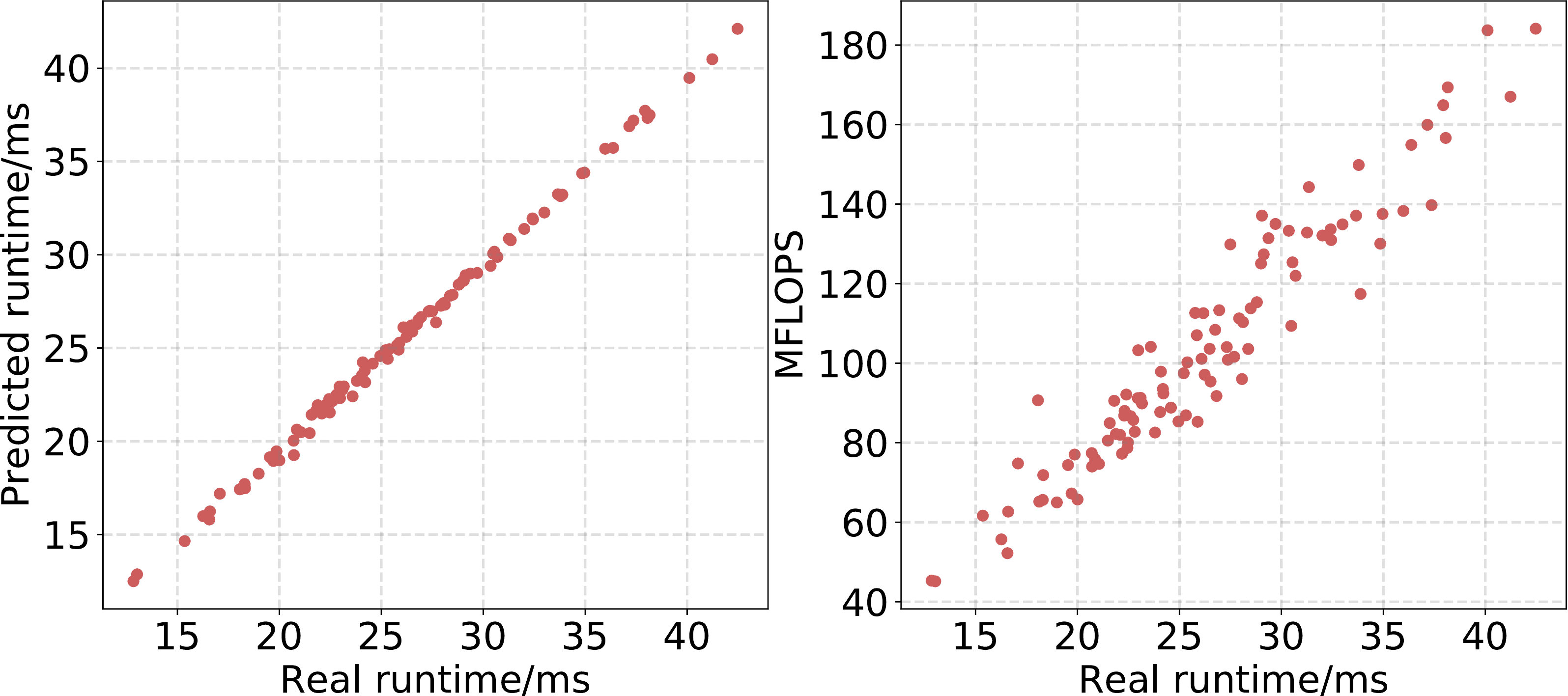}
    \end{center}
    \caption{Left: the real runtime \& the predicted runtime with proposed trilinear interpolation. Right: the real runtime \& the number of MACs (Multiply-ACcumulations) of models.}
    \label{fig:latency-estimation}
\end{figure}
To evaluate the accuracy of the proposed latency estimation, we compare the real latency and predicted latency of 100 randomly generated MobileNetV1 models with various number of channels and weight sparsity. Specifically, we run each model for 20 times, calculate the average runtime and compare it with the predicted runtime. Results are shown in the left of \Cref{fig:latency-estimation}. From the figure, we can observe that with trilinear interpolation, the predicted latency is highly correlated to the real latency of deep networks.

The right of \Cref{fig:latency-estimation} shows the latency and the number pf arithmetic operations of different deep networks. We can see that the number of arithmetic operations (MFLOPS) is positively correlated with latency generally, while the latency does not monotonically increase with the number of operations (MFLOPS). This is mainly because that the model's latency on real hardware platform can be impacted by both computation intensity and other factors such as memory access time. This phenomenon motivates us to design deep networks based on latency instead of FLOPs for model pruning. When deploying a deep network into a practical hardware, we consider the latency of runtime, not the FLOPs, for the deep network.

\noindent\textbf{Effect of joint channel and weight pruning.}
\newcommand{\yes}{\ding{51}}
\newcommand{\no}{\ding{55}}
\begin{table}[!bt]
    \centering
    \begin{tabular}{l|ccccc}
    \Xhline{2\arrayrulewidth}
        Method & CP & WP & Optim. & Latency & Accuracy \\
        \hline
        \multirow{4}{*}{WSO} & \multirow{4}{*}{\no} & \multirow{4}{*}{\yes} & \multirow{4}{*}{-} & 161.72 ms & 68.45\% \\
        & & & & 196.82 ms & 69.54\% \\
        & & & & 200.50 ms & 69.57\% \\
        & & & & 220.07 ms & 70.09\% \\
        \hline
        \multirow{4}{*}{CWO} & \multirow{4}{*}{\yes} & \multirow{4}{*}{\no} & \multirow{4}{*}{-} & 161.14 ms & 66.78\% \\
        & & & & 205.03 ms & 67.75\% \\
        & & & & 265.57 ms & 68.36\% \\
        & & & & 339.52 ms & 69.79\% \\
        \hline
        \multirow{4}{*}{SCW} & \multirow{4}{*}{\yes} & \multirow{4}{*}{\yes} & \multirow{4}{*}{Seq.} & 197.62 ms & 69.29\% \\
        & & & & 221.65 ms & 69.58\% \\
        & & & & 266.59 ms & 69.98\% \\
        & & & & 305.78 ms & 70.20\% \\
        \hline
        \rowcolor{cyan!25}
        & & & & \textbf{160.37 ms} & \textbf{69.16\%} \\
        \rowcolor{cyan!25}
        & & & & \textbf{193.65 ms} & \textbf{69.70\%} \\
        \rowcolor{cyan!25}
        & & & & \textbf{196.44 ms} & \textbf{69.90\%} \\
        \rowcolor{cyan!25}
        \multirow{-4}{*}{\textbf{JCW}} & \multirow{-4}{*}{\yes} & \multirow{-4}{*}{\yes} & \multirow{-4}{*}{\textbf{Joint}} & \textbf{223.73 ms} & \textbf{70.13\%} \\
    \Xhline{2\arrayrulewidth}
    \end{tabular}
    \caption{Comparisons among different variants of JCW for accelerating ResNet18 on ImageNet. CP, WP denote channel pruning and weight pruning, respectively. Seq. (Joint) denotes optimizing the layer-wise number of channels and weight sparsity sequentially (in one single optimization run).}
    \label{tab:joint-influence}
\end{table}
To study the effect of joint channel and weight pruning, we compare JCW with three of its variants, \textit{i.e.} (i) WSO which only searches for weight sparsity, while the number of channels for each layer remains the maximum value; (ii) CWO which only searches for the number of channels while keeps the full parameters of remaining channels; (iii) SCW which searches for number of channels and weight sparsity sequentially.

\Cref{tab:joint-influence} shows that our JCW outperforms all the other variants. In particular, under the latency of $\sim 160ms$, the accuracy of JCW is $0.61\%$ and $2.38\%$ higher than WSO and CWO, respectively. Moreover, under the similar accuracy of $\sim 70\%$, the latency of JCW is $1.37\times$ faster than SCW ($305.78ms \rightarrow 223.73ms$).

\noindent\textbf{Effect of uniform non-dominated sorting selection.}
\begin{figure}[t]
    \centering
    \includegraphics[width=\linewidth]{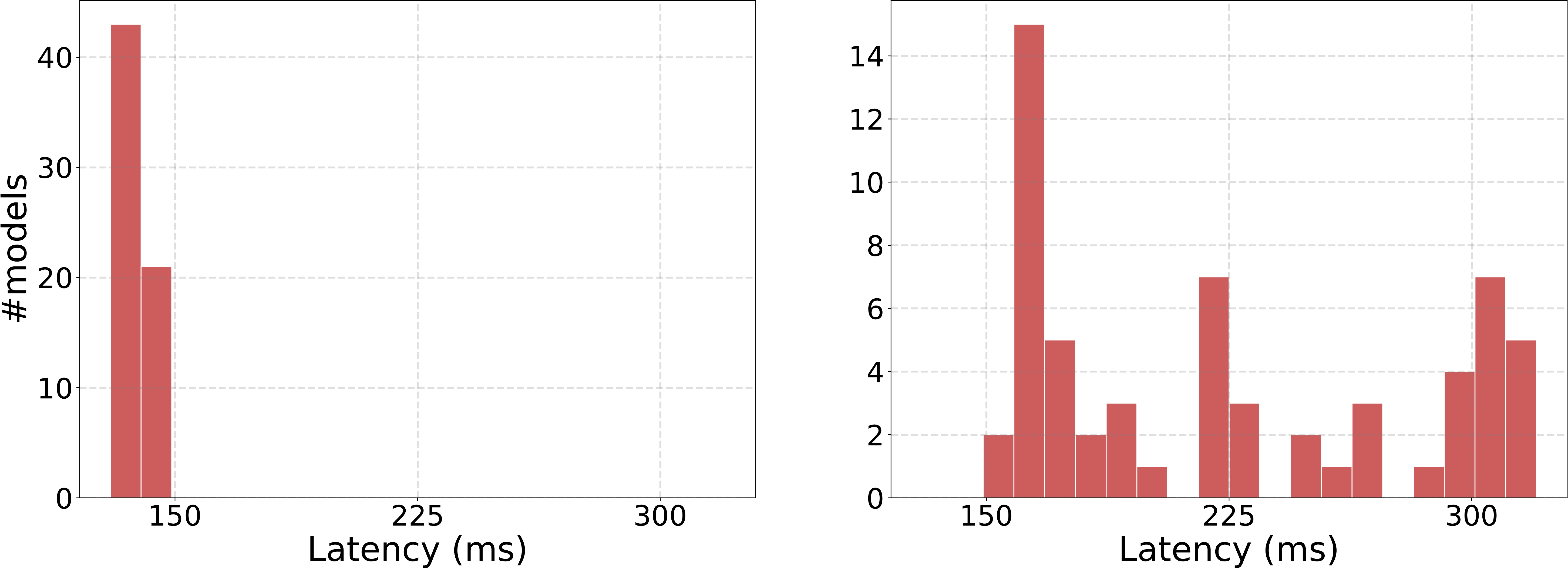}
    \caption{{Left: latency distribution after 50 evolutionary steps with standard non-dominated sorting selection. Right: latency distribution after 50 evolutionary steps with uniform non-dominated sorting selection. The searches are conducted with the ResNet18 on ImageNet. The proposed selection scheme generates models of various latency, while the standard selection gives priority to models of low latency.}}
    \label{fig:latency-dist}
\end{figure}
Considering the search efficiency, we predict the accuracy of models with different architectures using parameter sharing at the cost of inaccurate accuracy prediction. Because of inaccurate model accuracy prediction, the evolver tends to focus on minimizing the other objective, the latency. For instance, let $\mathcal{A}$, $\mathcal{B}$ be two architectures in the combined population to be selected. We assume that $\mathcal{T(A)}<\mathcal{T(B)}$, $\mathcal{E(A)}>\mathcal{E(B)}$. Here, $\mathcal{T}$ and $\mathcal{E}$ are real latency and error rate of architectures, respectively. In term of multi-objective optimization, there is no priority relation between $\mathcal{A}$ and $\mathcal{B}$. In other words, both architectures $\mathcal{A}$ and $\mathcal{B}$ should be selected in the new population with the equal chance. If the accuracy estimation is inaccurate, it is likely that the predicted error rate of $\mathcal{A}$ is smaller than $\mathcal{B}$. In this case, the standard non-dominated sorting selection may remove the architecture $\mathcal{B}$ from population incorrectly. This motivates us to develop the uniform non-dominated sort scheme, which explicitly selects architectures with a wide range of latency. The uniform non-dominated sorting selection generates diverse architectures of various latency.

To further validate the effectiveness of the proposed selection method, we show the latency distribution of models searched with the above two different selection methods in \Cref{fig:latency-dist}. The experiments are conducted with ResNet18 on the ImageNet dataset. From the left of \Cref{fig:latency-dist}, we can observe that the latency of models searched with the original selection scheme are small after 50 evolutionary steps. In contrast, from the right of \Cref{fig:latency-dist} we can observe that with the proposed uniform non-dominated sorting selection, models with relatively large latency are also preserved during search. So we conclude that the enhanced evolutionary algorithm with the proposed uniform non-dominated sorting selection is able to generate diverse models with various latency.

\subsection{Comparison with state-of-the-art methods}
\begin{table}[!bt]
    \centering
    \small
    \begin{tabular}{l|cccc}
        \Xhline{2\arrayrulewidth}
         Model & Method & Latency & Acc@1 & Acc@1$\uparrow$ \\
        \Xhline{1.5\arrayrulewidth}
         \multirow{7}{*}{ResNet18} & Uniform $1\times$ & 537 ms & 69.8\% & - \\
                                 ~ & DMCP~ & 341 ms & 69.7\% & -0.1\% \\
                                 ~ & APS~ & 363 ms & 70.2\% & +0.4\% \\
                                 \cline{2-5}
                                 \cline{2-5}
                                 \cline{2-5}
                                 ~ & \cellcolor{cyan!25}~ & \cellcolor{cyan!25}\textbf{160 ms} & \cellcolor{cyan!25}\textbf{69.2\%} & \cellcolor{cyan!25}\textbf{-0.6\%} \\
                                 ~ & \cellcolor{cyan!25}~ & \cellcolor{cyan!25}\textbf{194 ms} & \cellcolor{cyan!25}\textbf{69.7\%} & \cellcolor{cyan!25}\textbf{-0.1\%} \\
                                 ~ & \cellcolor{cyan!25}~ & \cellcolor{cyan!25}\textbf{196 ms} & \cellcolor{cyan!25}\textbf{69.9\%} & \cellcolor{cyan!25}\textbf{+0.1\%} \\
                                 ~ & \cellcolor{cyan!25}\multirow{-4}{*}{\textbf{JCW (OURS)}} & \cellcolor{cyan!25}\textbf{224 ms} & \cellcolor{cyan!25}\textbf{70.2\%} & \cellcolor{cyan!25}\textbf{+0.4\%} \\
         \hline
         \multirow{14}{*}{MobileNetV1} & Uniform $1\times$ & 167 ms & 70.9\% & - \\
                                   ~ & Uniform $0.75\times$ & 102 ms & 68.4\% & - \\
                                   ~ & Uniform $0.5\times$ & 53 ms & 64.4\% & - \\
                                   ~ & AMC  & 94 ms & 70.7\% & -0.2\% \\
                                   ~ & Fast $^\dag$ & 61 ms & 68.4\% & -2.5\% \\
                                   ~ & AutoSlim & 99 ms & 71.5\% & +0.6\% \\
                                   ~ & AutoSlim  & 55 ms & 67.9\% & -3.0\% \\
                                   ~ & USNet  & 102 ms & 69.5\% & -1.4\% \\
                                   ~ & USNet  & 53 ms & 64.2\% & -6.7\% \\
                                   \cline{2-5}
                                   \cline{2-5}
                                   \cline{2-5}
                                   ~ & \cellcolor{cyan!25}~ & \cellcolor{cyan!25}\textbf{31 ms} & \cellcolor{cyan!25}\textbf{69.1\%} & \cellcolor{cyan!25}\textbf{-1.8\%} \\
                                   ~ & \cellcolor{cyan!25}~ & \cellcolor{cyan!25}\textbf{39 ms} & \cellcolor{cyan!25}\textbf{69.9\%} & \cellcolor{cyan!25}\textbf{-1.0\%} \\
                                   ~ & \cellcolor{cyan!25}~ & \cellcolor{cyan!25}\textbf{43 ms} & \cellcolor{cyan!25}\textbf{69.8\%} & \cellcolor{cyan!25}\textbf{-1.1\%} \\
                                   ~ & \cellcolor{cyan!25}~ & \cellcolor{cyan!25}\textbf{54 ms} & \cellcolor{cyan!25}\textbf{70.3\%} & \cellcolor{cyan!25}\textbf{-0.6\%} \\
                                   ~ & \cellcolor{cyan!25}\multirow{-5}{*}{\textbf{JCW (OURS)}} & \cellcolor{cyan!25}\textbf{69 ms} & \cellcolor{cyan!25}\textbf{71.4\%} & \cellcolor{cyan!25}\textbf{+0.5\%} \\
         \hline
         \multirow{14}{*}{MobileNetV2} & Uniform $1\times$ & 114 ms & 71.8\% & - \\
                                   ~ & Uniform $0.75\times$ & 71 ms & 69.8\% & - \\
                                   ~ & Uniform $0.5\times$ & 41 ms & 65.4\% & - \\
                                   ~ & APS & 110 ms & 72.8\% & +1.0\% \\
                                   ~ & APS &  64 ms & 69.0\% & -2.8\% \\
                                   ~ & DMCP & 83 ms & 72.4\% & +0.6\% \\
                                   ~ & DMCP & 45 ms & 67.0\% & -4.8\% \\
                                   ~ & DMCP & 43 ms & 66.1\% & -5.7\% \\
                                   ~ & Fast$^\dag$ & 89 ms & 72.0\% & +0.2\% \\
                                   ~ & Fast$^\dag$ & 62 ms & 70.2\% & -1.6\% \\
                                   \cline{2-5}
                                   \cline{2-5}
                                   \cline{2-5}
                                   ~ & \cellcolor{cyan!25}~ & \cellcolor{cyan!25}\textbf{30 ms} & \cellcolor{cyan!25}\textbf{69.1\%} & \cellcolor{cyan!25}\textbf{-2.7\%}\\
                                   ~ & \cellcolor{cyan!25}~ & \cellcolor{cyan!25}\textbf{40 ms} & \cellcolor{cyan!25}\textbf{69.9\%} & \cellcolor{cyan!25}\textbf{-1.9\%}\\
                                   ~ & \cellcolor{cyan!25}~ & \cellcolor{cyan!25}\textbf{44 ms} & \cellcolor{cyan!25}\textbf{70.8\%} & \cellcolor{cyan!25}\textbf{-1.0\%} \\
                                   ~ & \cellcolor{cyan!25}\multirow{-4}{*}{\textbf{JCW (OURS)}} & \cellcolor{cyan!25}\textbf{59 ms} & \cellcolor{cyan!25}\textbf{72.2\%} & \cellcolor{cyan!25}\textbf{+0.4\%} \\
         \Xhline{2\arrayrulewidth}
    \end{tabular}
    \caption{Comparison of JCW with various state-of-the-art approaches on the ImageNet dataset. The last column lists the accuracy improvement compared to the dense baseline \textit{i.e.} the uniform $1\times$ models. $^\dag$ means that the compared method is a random weight pruning method, and we measure the latency of their models with codes released by the authors \cite{google2020xnnpack}.}
    \label{tab:compare-with-sota}
\end{table}
We compare JCW with previous state-of-the-art channel pruning methods, DMCP~\citep{guo2020dmcp}, APS~\citep{wang2020revisiting}, AMC~\citep{he2018amc}, AutoSlim~\citep{yu2019autoslim}, USNet~\citep{yu2019universally}, and efficient weight pruning methods, fast sparse convolution~\citep{elsen2020sparsecov}, in terms of model latency and accuracy. Fast sparse convolution~\citep{elsen2020sparsecov} proposes an efficient algorithm for sparse convolution computation, and we measure the latency of their models with the released code~\citep{google2020xnnpack}. Results are shown in \Cref{tab:compare-with-sota}. We can observe that JCW consistently outperforms all the previous state-of-the-art methods in terms of model latency and accuracy by a large margin. Experimental results support our claim that the JCW framework improves the network pruning given a variety of latency budget.

\noindent\textbf{Results on ResNet18.} The top of \Cref{tab:compare-with-sota} shows the results of ResNet18 on the ImageNet dataset. Our method accelerates the inference of ResNet18 by $2.74\times$ without any accuracy loss. Compared to DMCP~\citep{guo2020dmcp}, JCW further accelerates the inference by $1.76\times$ with the same accuracy of $69.7\%$. The JCW has an acceleration ratio of $1.62\times$ over ASP~\citep{wang2020revisiting} with the same accuracy.

\noindent\textbf{Results on MobileNetV1.} From the middle of \Cref{tab:compare-with-sota}, we observe that JCW reduces the latency of MobileNetV1 by $2.42\times$ even with $0.5\%$ higher accuracy. With similar accuracy of $71.5\%$ and $71.4\%$, the latency of JCW is $1.43\times$ lower than that AutoSlim~\citep{yu2019autoslim}. Compared to fast sparse convolution~\citep{elsen2020sparsecov}, which is the currently state-of-the-art acceleration method for weight pruning, JCW achieves $1.9\%$ higher accuracy ($68.4\%\rightarrow 70.3\%$) with $1.13\times$ lower latency ($61$ ms $\rightarrow$ $54$ ms).

\noindent\textbf{Results on MobileNetV2.} The bottom of \Cref{tab:compare-with-sota} shows the results of MobileNetV2 on ImageNet. We can observe that, compared to the original MobileNetV2, JCW reduces the latency by $1.93\times$ even with $0.4\%$ higher accuracy.

To sum up, the performance of different models verifies that the superiority of our method is invariant to the types of the well-performed base architectures.

\section{Related Works}

Pruning has long been one of the primary techniques for network compression and acceleration~\citep{han2015lwc,han2016deepcompression,li2016pruning,he2017channelprune,liu2019metapruning}. These methods removes unimportant parameters from the original network and optimizes the remaining parts of the networks to retain accuracy. According to the granularity of pruning, these methods can be categorized into fine-grained weight pruning and coarse-grained filter pruning. In weight pruning, parameters are removed in weight-level~\citep{lecun1990optimal,han2015lwc,han2016deepcompression,ding2019gmsgd,frankle2019lottery}. Weight pruning is flexible to achieve theoretically smaller models and can also be efficiently accelerated on mobile CPUs thanks to the recent work of fast sparse convolution~\citep{elsen2020sparsecov}. In contrast, channel pruning compresses networks by removing parameters at the filter-level. Most of the early channel pruning methods are based on an filter importance scoring scheme, \textit{e.g.}, the filter norm~\citep{li2017peec,he2018sfp}, the percentage of zero-activation~\citep{hu2016apoz}, the reconstruction error of outputs~\citep{he2017channelprune,luo2018thinet,ding2019aofp}, the increase of loss after pruning~\citep{molchanov2017taylor-o1,molchanov2019taylor-o1,peng2019collaborative,liu2021groupfisher}, the geometric properties of filters~\citep{he2019fpgm,joo2021linearly}. Besides, sparse regularization based methods~\citep{wenwei2016SSL,ruan2021dpfps} have also been intensively explored. Channel pruning methods are well-structured, thus it can be directly accelerated without extra implementation efforts. 


Apart from the aforementioned pruning methods, many recently emerging pruning methods formulate the network pruning as an architecture search, taking benefits from the automated process in composing architectures to avoid the labor-prohibitive model design, \textit{e.g.}, determination of layer-wise channel numbers for channel pruning. \citet{he2018amc} propose to determine the number of channels with reinforcement learning and outperforms human-designed pruning methods. \citet{lin2020abc} search for the channel numbers with population-based algorithm. \citet{wang2021srr} model the problem of channel number search as structural redundancy reduction. \citet{wang2021comprehensive,gao2021nppm} train parameterized accuracy predictors to guide the pruning process. \citet{liu19metapruning} train a meta network to predict the weights of compressed models, then conduct evolutionary search for pruning. There is also a vast body of work utilizing the parameter sharing technique to train a supernet for accuracy evaluation and conduct the pruning with evolutionary search~\citep{guo2020single,cai2020once}, greedy slimming~\citep{yu2019autoslim}, or reinforcement learning~\citep{wang2020revisiting}. Besides, differentiable channel number search approaches~\citep{dong2019tas,guo2020dmcp} have also been investigated.


Efficient model design often involves multiple objectives, \textit{e.g.}, the accuracy, the latency, the energy, the model size, et cetera. In this perspective, it is more desired to search for a sequence of Pareto optimal models. Many works have been proposed to deal with multi-objective model design. \citet{tan2019mnasnet,hsu2018monas} integrate multiple objectives into one correlated reward function and conduct the search with reinforcement learning. However, they need trail and error to design the form and related hyper-parameters for the correlated reward function, which requires expert knowledge and is laborious. Some recent works~\citep{dong2018dpp,lu2019nsga,elsken2019efficient} search for Pareto optimal architectures directly with evolutionary algorithm and a selection criterion based on non-dominated sorting~\citep{deb2002nsga2}. Our work focuses on model pruning, which is orthogonal to these general NAS methods.

The main difference of JCW is to investigate the essential part of network pruning: is it possible to absorb both benefits of channel and weight pruning and achieve a better accuracy-latency trade-off by applying the two jointly?  We conduct extensive experiments and ablation studies, which demonstrate that the joint channel and weight pruning achieves a better accuracy-latency Pareto frontier than previous network pruning approaches.



\section{Conclusion}
In this work, we propose a joint pruning for both channel and weight, named JCW. Channel pruning provides instant acceleration while weight pruning are more flexible to preserve model accuracy. We construct a multi-objective optimization considering both model accuracy and inference latency in the JCW, which can be solved by a tailored Pareto-optimization evolutionary algorithm. Extensive experiments demonstrate that the effectiveness of each component of JCW. The JCW outperforms previous state-of-the-art model compression and acceleration approaches with ResNet18, MobileNetV1 and MobileNetV2 on the ImageNet classification dataset. 



%

{
\bibliographystyle{plainnat}

\bibliography{sample-base}}

\newpage
\appendix
\section{Details of evolutionary search}
\subsection{Feature design}
In this section, we introduce the feature design used during evolutionary search. Denote $C_{\max}=\{c_{\max}^{(l)}\}_{l=1}^L$ as the maximum number of channels for each layer. Then for architecture $(C, S)$ with number of channels $C$ (for channel pruning) and layer-wise weight sparsity (for weight pruning), the feature of $(C, S)$ can be denoted by:
\begin{equation}
    f(C, S) = (\frac{C}{C_{\max}}, S)
\end{equation}
In this way, all the dimensions of features are normalized to $[0, 1]$. It is straight forward to re-generate architectures from features:
\begin{equation}
    f^{(-1)}(\tilde{C}, S) = (\lfloor\tilde{C}\times C_{\max}\rceil, S)
\end{equation}

\subsection{Crossover and mutation}
In this section, we introduce details about the crossover and mutation operations used in our method.

\textbf{Crossover.} For crossover operation, we use the popular Simulated Binary Crossover operation (SBX). We start with the crossover operation between two scalar numbers $\alpha, \beta$. After the crossover operation, two new individuals are generated:
\begin{equation}
    \begin{split}
        & \alpha^{new} = 0.5\times [(1+u)\alpha + (1-u)\beta] \\
        & \beta^{new} = 0.5\times [(1-u)\alpha + (1+u)\beta]
    \end{split}
\end{equation}
where:
\begin{equation}
    u = \begin{cases}
          (2p)^{\frac{1}{\mu_c + 1}} ~~~~ & ~~~~ p\le 0.5 \\
          (\frac{1}{2(1-p)})^{\frac{1}{\mu_c + c}}        &      otherwise
        \end{cases}
\end{equation}
where $p$ is a random number sampled from $[0, 1]$, and $\mu_c$ is a hyper-parameter that makes a trade-off between exploration and exploitation. For two $n-dim$ vectors $\boldsymbol{\alpha} = \{\alpha_i\}_{i = 1}^n$ and $\boldsymbol{\beta} = \{\beta_i\}_{i=1}^{(n)}$, the crossover operation between $\boldsymbol{\alpha}$ and $\boldsymbol{\beta}$ can be implemented by applying the crossover operation on each dimension of the two vectors:
\begin{equation}
    \begin{split}
        & \boldsymbol{\alpha}^{new} = \{\alpha_i^{new}\}_{i = 1}^n \\
        & \boldsymbol{\beta}^{new} = \{\beta_i^{new}\}_{i = 1}^n
    \end{split}
\end{equation}
where $\alpha_i^{new}$ and $\beta_i^{new}$ are the output of crossover operations between $\alpha_i$ and $\beta_i$. For ease of explanation, we denote:
\begin{equation}
    \{\boldsymbol{\alpha}^{new}, \boldsymbol{\beta}^{new}\} = SBX(\boldsymbol{\alpha, \beta})
\end{equation}
With the above operation, the crossover operation on a population $P$ can be implemented with \Cref{alg:crossover}.

\begin{algorithm}[!t]
  \centering
  \caption{Crossover}
  \label{alg:crossover}
  \begin{algorithmic}[1]
    \REQUIRE{The population $P$ with size $n$.}
    \ENSURE{The population $Q$ with size $n$.}
    \STATE {$Q = \emptyset$}
    \WHILE{$|Q|<n$}
      \STATE{Randomly sample two individuals $p_1, p_2$ from $P$.}
      \STATE{$Q = Q\cup SBX(p_1, p_2)$}
    \ENDWHILE
    \RETURN {$Q$}
  \end{algorithmic}
\end{algorithm}

\textbf{Mutation.} For mutation operation, we use the popular PoLynomial Mutation operation (PLM) for MOO genetic algorithms. We denote $u_i, l_i$ as the upper bound and lower bound of the $i^{th}$ dimension of individuals. The algorithm for mutation are summarized in \Cref{alg:mutation}, where $\mu_m$ is a hyper-parameter.
\begin{algorithm}
\small
  \centering
  \caption{Mutation}
  \label{alg:mutation}
  \begin{algorithmic}[1]
    \REQUIRE{A population $P$ with size $n$, and feature dimension $d$ for each individual. The mutation probability $p$.}
    \ENSURE{A mutated population $Q$.}
    \STATE{$Q = \emptyset$}
    \FOR {each individual $p\in P$}
      \STATE{$q = p$}
      \FOR {each dimension $i\in1\rightarrow d$}
        \STATE {$\mu = $ randomly sample from $[0, 1]$.}
        \IF {$\mu<p$}
          \STATE {$d_1 = \frac{p_i - l_i}{u_i - l_i}, d_2 = \frac{u_i - p_i}{u_i - l_i}$}
          \STATE {$r = $ randomly sample from $[0, 1]$.}
          \STATE {$q_i = \begin{cases}
                          & [2r + (1 - 2r)(1 - d_1)^{\mu_m + 1}]^{\frac{1}{\mu_m + 1}}  ~ r \le 0.5 \\
                          & 1 - [2(1-r) + 2(r-0.5)(1-d_2)^{\mu_m + 1}]^{\frac{1}{\mu_m + 1}}
                         \end{cases}$}
        \ENDIF
      \ENDFOR
      \STATE{$Q = Q\cup \{q\}$.}
    \ENDFOR
    \RETURN {$Q$}
  \end{algorithmic}
\end{algorithm}

\textbf{Crossover mutation in our method.} Having introduced the crossover and mutation operations, we are able to introduce the final crossover-mutation operations used in our method. The algorithm is summarized in \Cref{alg:crossover-mutation}. In all of our algorithm, we set $\mu_c = 0.5$, $\mu_m = 15$, and the probability for mutation to 0.1.

\begin{algorithm}[!tbh]
    \centering
    \caption{Crossover and mutation}
    \label{alg:crossover-mutation}
    \begin{algorithmic}
      \REQUIRE{A set of architectures $P = \{(C_i, S_i)\}_{i=1}^{n}$ with population size $n$.}
      \ENSURE{A set of architectures $Q$ generated from $P$.}
      \STATE {$\tilde{P} = \{f(C_i, S_i)\}_{i = 1}^{n}$}
      \STATE {$\tilde{Q}$ = crossover between individuals in $P$ with \Cref{alg:crossover}.}
      \STATE {Clip the values of $\tilde{Q}$ to $[0, 1]$.}
      \STATE {$\tilde{Q}$ = mutate $\tilde{Q}$ with \Cref{alg:mutation}.}
      \STATE{Clip the values of $\tilde{Q}$ to $[0, 1]$.}
      \STATE{$Q = \{f^{(-1)}(\tilde{C}_i, S_i)\}_{(\tilde{C}_i, S_i)\in\tilde{Q}}$}
      \RETURN {$Q$}
    \end{algorithmic}
\end{algorithm}

\begin{algorithm}[!tbh]
  \centering
  \caption{JCW algorithm}
  \label{alg:jcw}

  \begin{algorithmic}[1]
    \REQUIRE{A supernet $\mathcal{N}$}, the number of generations $N_G$, the population size $n$.
    \ENSURE{A set of architectures $P = \{(C_1, S_1), \cdots, (C_n, S_n)\}$ with various latency and accuracy.}
    \STATE{$P$ = Randomly sample $n$ architectures with various number of channels and weight sparsity for each layer.}
    \FOR{$g\in 1\rightarrow N_G$}
      \STATE{$Q=crossover\_mutation(P)$ with \Cref{alg:crossover-mutation}}
      \STATE{Estimate the accuracy of models in $P\cup Q$ with parameter sharing as described in section 2.4.}
      \STATE{Estimate the latency of models in $P\cup Q$ with trilinear interpolation as described in section 2.5.}
      \STATE{Update $P$ with uniform non-dominated sorting selection as described in Section 2.6.}
    \ENDFOR
    \STATE{Train the models with architectural configurations in $P$ with ADMM.}
  \end{algorithmic}
\end{algorithm}

\section{Model retraining}
After searching, we get the optimal layer-wise channel width as well as the weight sparsity. The generated models are retrained on the whole train set of ImageNet and validated on validation set. We follow the conventional pretrain + prune + fine tune process for model retraining. Specifically, for an architecture $(C, S)$, we first train a dense network with number of channels $C$. We then conduct the weight pruning with ADMM. By introducing an auxiliary variable and using the duality theorem, the primal parameters $W$, auxiliary variable $U$ and the dual variable $Z$ are updated alternatively:
\begin{equation}
    \begin{split}
       & W_{t+1} = \arg\min_{W} \mathcal{L}(W) + \frac{\rho}{2}\|W - U_t + Z_t\|^2 \\
       & U_{t+1} = \arg\min_{U} \|W_{t+1} - U + Z_t\|^2~~~~s.t.~~~~\|U^{(l)}\|_0 = s^{(l)} \\
       & Z_{t+1} = Z_t + (W_{t+1} - U_{t+1})
    \end{split}
\end{equation}
where $\mathcal{L}(W)$ is task specified loss function. $U$ and $Z$ are all of the same size as $W$, and $U^{(l)}$ is the auxiliary variable corresponding to parameters of the $l^{th}$ layer. After ADMM steps, we fine tune the compressed model for 60 epochs. Detailed hyper parameters for model retraining are listed in \Cref{tab:hyper-param-retrain}, and the final algorithm of JCW is summarized in \Cref{alg:jcw}.

\begin{table*}[!tbh]
    \centering
    \begin{tabular}{l|c|ccc}
        \toprule
        \hline
                                Stage &          Hyper parameter & Resnet18 & MobileNetV1 & MobileNetV2 \\
        \hline
        \multirow{5}{*}{Pretrain} & batch size & 512 & 512 & 512 \\
                               ~ & epochs  & 120 & 120 & 250 \\
                               ~ & lr & 0.256 & 0.512 & 0.256 \\
                               ~ & lr annealing & cosine & cosine & cosine \\
                               ~ & weight decay & 1e-4 & 4e-5 & 4e-5 \\
        \hline
        \multirow{6}{*}{ADMM} & batch size & 512 & 512 & 512 \\
                           ~ & epochs & 60 & 60 & 60 \\
                           ~ & lr & 0.005 & 0.005 & 0.005 \\
                           ~ & lr annealing & constant & constant & constant \\
                           ~ & weight decay & 1e-4 & 4e-5 & 4e-5 \\
                           ~ & $\rho$ & 0.01 & 0.01 & 0.01 \\
        \hline
        \multirow{5}{*}{Fine tune} & batch size & 512 & 512 & 512 \\
                                ~ & epochs & 60 & 60 & 60 \\
                                ~ & lr & 0.005 & 0.005 & 0.005 \\
                                ~ & lr annealing & cosine & cosine & cosine \\
                                ~ & weight decay & 0 & 0 & 0\\
        \hline
        \bottomrule
    \end{tabular}
    \caption{Hyper-parameters for model re-training.}
    \label{tab:hyper-param-retrain}
\end{table*}

\section{Linearity of latency w.r.t. sparsity}
\begin{figure}[!tbh]
    \centering
    \includegraphics[width=\linewidth]{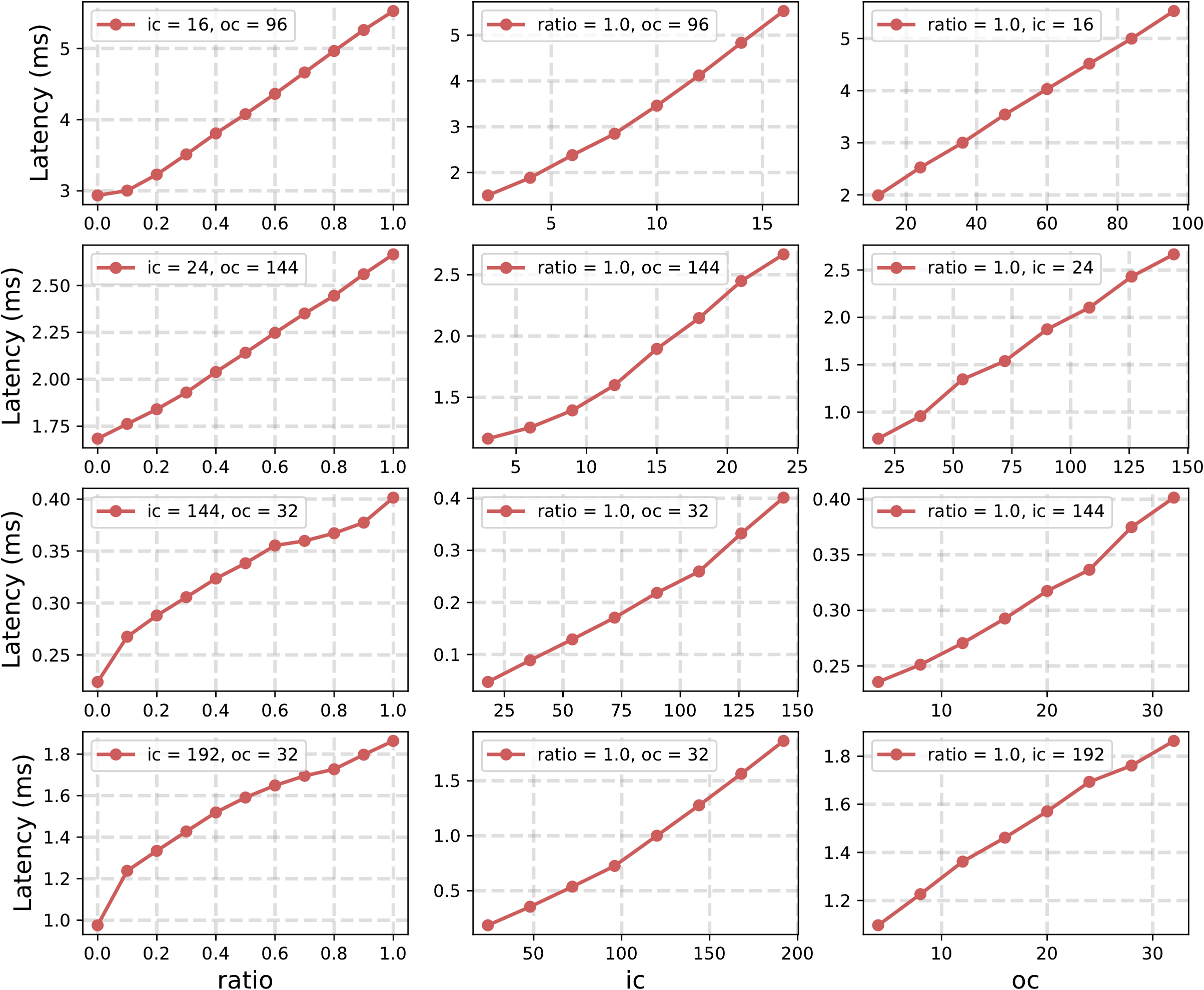}
    \caption{Linearity of latency w.r.t. density ratio (left), input channels (middle) and output channels (right).}
    \label{fig:linearity}
\end{figure}

In \Cref{fig:linearity} we plot the latency v.s. the number of input channels, the number of output channels and the weight sparsity for weight pruning. The data points are collected from 4 of the convolution layers of MobileNetV2. We can see that the latency of each layer is locally linear to the layer width and weight sparsity, this motivates us to approximate the latency of networks with tri-linear interpolation.

\noindent\textbf{Derivation of tri-linear interpolation.}
We further derive the tri-linear interpolation of Equation (9) in Section 2.5 of the paper.

\begin{figure}[!tb]
    \centering
    \includegraphics[width=0.6\linewidth]{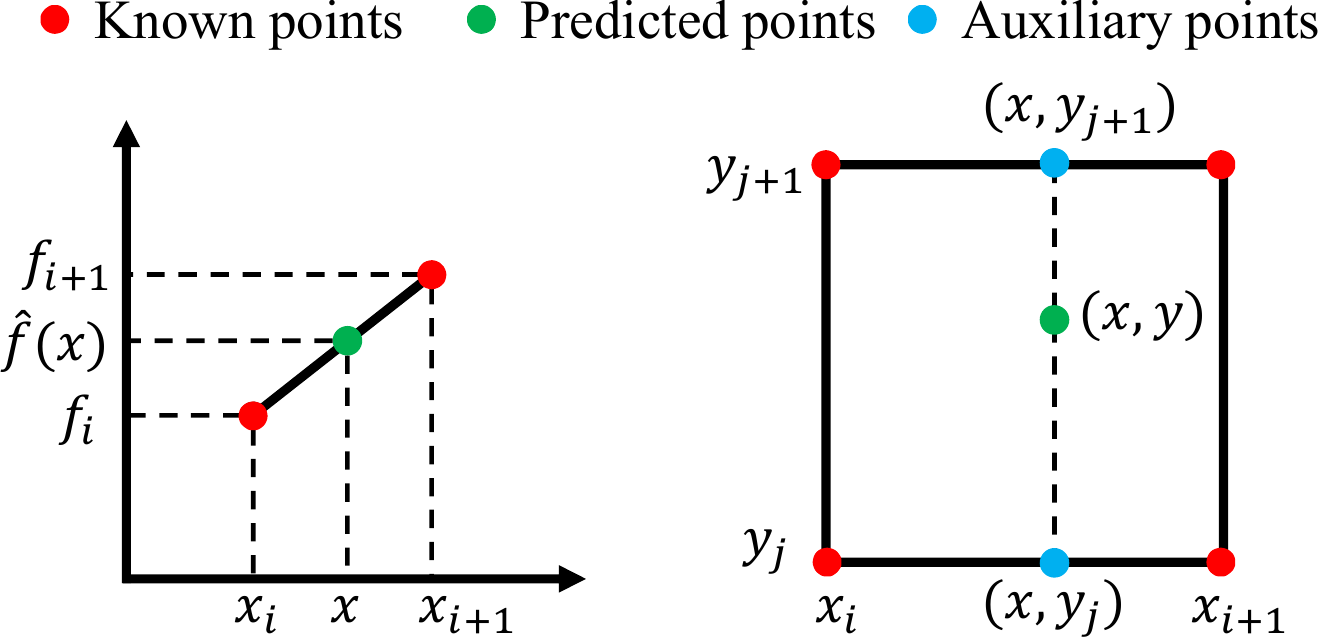}
    \caption{An illustration of \textit{left:} 1-d linear interpolation and \textit{right:} 2-d bi-linear interpolation.}
    \label{fig:illustration-linear-interp}
\end{figure}

\noindent{\textbf{1-d linear interpolation.}} We start from 1-d linear interpolation, which is illustrated in the left of \Cref{fig:illustration-linear-interp}. Assume that we hope to approximate some function $f(\cdot)$ with an array of known data points $\{x_i, f_i\}_{i=0}^{n}$, where:
\begin{equation*}
    x_i = \frac{i}{n}x_n, ~~~~ f_i = f(x_i),
    \label{eq:def-x-f}
\end{equation*}
and $x_n$ is the maximum value of $x$. Denote $\hat{f}(\cdot)$ to be the approximation to $f(\cdot)$ with linear interpolation in the sequence of known data points. Given any position $x$ in the interval $[x_i, x_{i+1}]$, the approximated function $\hat{f}(x)$ is the straight line between the pair of data points $(x_i, f_i)$ and $(x_{i+1}, f_{i+1})$, we then have:
\begin{equation*}
    \frac{f_{i+1} - f_i}{x_{i+1} - x_i} = \frac{\hat{f}(x) - f_i}{x - x_i}.
\end{equation*}
Solving the above linear equation, we have:
\begin{equation}
\begin{split}
    \hat{f}(x) &= f_i\times\left(1 - |\frac{x - x_i}{\triangle_x}|\right)  \\
               &+ f_{i+1}\times\left(1 - |\frac{x - x_{i+1}}{\triangle_x}|\right)
\end{split},
    \label{eq:2-term}
\end{equation}
where:
\begin{equation*}
    \triangle_x = x_{i+1} - x_i = \frac{x_n}{n}.
    \label{eq:def-delta}
\end{equation*}
Note that for any $j < i$, we have:
\begin{equation*}
    |x - x_j| = (x - x_i) + (x_i - x_j) \ge \triangle_x,
\end{equation*}
similarly, for any $j > i + 1$, we also have:
\begin{equation*}
    |x - x_j| \ge \triangle_x.
\end{equation*}
Thus, \Cref{eq:2-term} can be further reorgnized by:
\begin{equation*}
\begin{split}
    \hat{f}(x) & = \sum_{i=0}^n \tau(\frac{x - x_i}{\triangle_x})f_i \\
               & = \sum_{i=0}^n\tau(n\frac{x}{x_n} - i)f_i
\end{split},
    \label{eq:sum}
\end{equation*}
where:
\begin{equation*}
    \tau(x) = \max(0, 1 - |x|).
\end{equation*}

\noindent{\textbf{Multi dimensional linear interpolation.}} The linear interpolation in higher-dimensional spaces can be done by conducting linear interpolation along each dimension separately. In the right of \Cref{fig:illustration-linear-interp}, we show a simple example for 2-d linear interpolation, or bi-linear interpolation. Specifically, in 2-d case, our goal is to approximate the values of some function $f:\mathbb{R}^2\rightarrow \mathbb{R}$ with bi-linear interpolation given a grid of known data points $\{(x_i, y_j, f_{ij}); i = 0,1\cdots n, j = 0, 1\cdots m\}$, where:
\begin{equation*}
    x_i = \frac{i}{n}x_{n}, ~~ y_j = \frac{j}{m}y_{m}, ~~ f_{ij} = f(x_i, y_j).
\end{equation*}
Given any point $(x, y)$ such that:
\begin{equation*}
    x\in[x_i, x_{i+1}], ~~ y\in[y_j, y_{j+1}],
\end{equation*}
the function value $f(x, y)$ can be then approximated in two steps. First, conduct the 1-d linear interpolation along the $x$-dimension, which gives:
\begin{equation*}
    \hat{f}(x, y_j) = \sum_{i} \tau(n\frac{x}{x_{n}} - i)f_{ij},
\end{equation*}
and then conduct the 1-d linear interpolation along the $y$-dimension, which further gives:
\begin{equation*}
\begin{split}
    \hat{f}(x, y) & = \sum_j \tau(m\frac{y}{y_{m}} - j)\hat{f}(x, y_j) \\
                  & = \sum_{i,j}\tau(n\frac{x}{x_{n}} - i)\tau(m\frac{y}{y_{m}} - j)f_{ij}
\end{split}.
\end{equation*}
The above derivation can be easily generalized to higher dimensional spaces. Particularly, in 3-d case, the tri-linear interpolation has the form as illustrated in Equation (9) of our paper.


\end{document}